\documentclass[conference]{IEEEtran}
\IEEEoverridecommandlockouts

\usepackage{cite}
\usepackage{amsmath,amssymb,amsfonts}
\usepackage{algorithmic}
\usepackage{graphicx}
\usepackage{textcomp}
\usepackage{xcolor}
\usepackage{xcolor}
\usepackage{booktabs} 
\usepackage{multirow}
\usepackage{url}
\def\BibTeX{{\rm B\kern-.05em{\sc i\kern-.025em b}\kern-.08em
    T\kern-.1667em\lower.7ex\hbox{E}\kern-.125emX}}
\begin{document}

\title{Is My Vision-Language Data in Your AI? \\Membership Inference Test (MINT) Demo 2}


\author{
    \IEEEauthorblockN{Daniel DeAlcala\IEEEauthorrefmark{1}, Gonzalo Mancera\IEEEauthorrefmark{1}, Julian Fierrez, Aythami Morales, Ruben Tolosana, Ruben Vera-Rodriguez}
    \IEEEauthorblockA{\textit{BiometricsAI, Universidad Autónoma de Madrid, Spain} \\
    \{daniel.dealcala, gonzalo.mancera, julian.fierrez, aythami.morales,  ruben.tolosana, ruben.vera\}@uam.es}
       \IEEEauthorblockA{\IEEEauthorrefmark{1}(Equal contribution)}
}

\maketitle

\begin{abstract}
We present the Membership Inference Test (MINT) Demo 2, a framework designed to improve transparency in machine learning training processes. MINT is a technique for experimentally determining whether specific data were used during machine learning model training. We establish the theoretical framework and propose multiple architectures for MINT depending on the amount of information known about the models that are being audited. Experimental results using a popular face recognition model, 4 state-of-the-art LLMs, and multiple, diverse, and large-scale public image and text databases achieve promising accuracy levels in the detection of training data of up to 90\%. Building on these results, we introduce a comprehensive web platform\footnote{\label{note:plataforma}\url{https://ai-mintest.org/}} that expands these capabilities to image and text modalities. The platform integrates a diverse technological stack, including MINT, aMINT, and gMINT, allowing users to audit a wide range of models. This demonstrator aims to promote AI transparency and provides a practical tool to foster compliance with emerging AI regulations.

\end{abstract}


\section{Introduction}
\label{sec:intro}

On 20 June 2023, the European Parliament adopted its negotiating position on the Artificial Intelligence Act \cite{AIact}. The parliament imposes an obligation on providers of AI technologies to ensure robust protection of fundamental citizen rights. This new regulation imposes the registration of AI models in an EU database and gives national authorities the power to request access to both the trained and training models of AI systems. This new regulation is a game changer that imposes transparency as a must to deploy AI technologies in Europe. As a result of this AI Act, among other relevant factors \cite{2026irigoyen-risks}, there is a growing need for new auditing tools to monitor AI technologies and their secure deployment in our society. 

Unauthorized use of personal or copyrighted material for training AI models can infringe on the rights of citizens. Moreover, the generated output of AI models trained on these data might blur the lines between original and derived works, leading to issues of plagiarism and copyright infringement. These considerations lead us to the main objective of this work, which is to propose a platform to detect the data used to train AI models. Currently, developers can hide behind the weights of their network to bypass regulations and hide the use of training data from users. employing the ideas presented in this work, it may be possible to unveil the AI model training process and align this training with current legislation and citizen rights \cite{2023human-centric,pena25llms}.

\begin{figure}[t]
\centering
\includegraphics[width=0.85\linewidth]{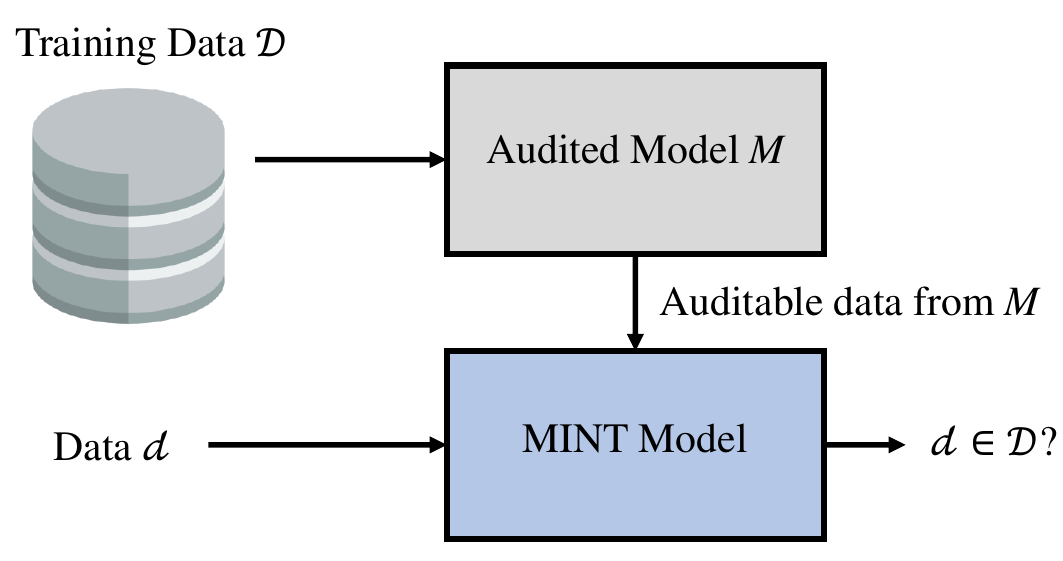} 
\caption{Block diagram of the Membership Inference Test (MINT).} 
\label{Block_diagram_full}
\end{figure}

Membership Inference Attacks (MIA) are privacy attacks that target machine learning models trained on sensitive data \cite{shokri2017membership,carlini2022membership}. The purpose of these attacks is to infer sensitive information from target models \cite{2021_TPAMI_SensitiveNets_Morales}. We propose a reformulation of AI privacy analysis methods \cite{2017_Access_HEmultiDTW_Marta,2022_Access_DP-CL_Ahmad} to be used as an auditing tool to detect possible data usage without owners' consent, which we refer to as the Membership Inference Test (MINT), as summarized in Fig.~\ref{Block_diagram_full}. The main contributions of this work can be summarized as follows:

\begin{itemize}
    \item We introduce MINT, a method for detecting the utilization of specific data during the training of AI models.
    \item We present experiments on $6$ diverse image databases with more than $22$M images and $6$ large-scale text datasets using 4 state-of-the-art LLMs. Accuracy results are obtained of up to 90\% in detecting training data in this challenging experimental setup.
    \item We introduce an upgraded interactive web platform as a demonstrator. This new MINT Demo 2 expands the technology from the previous MINT Demo \cite{dealcala2025mintdemo} to a multimodal framework, supporting both image and text models through a diverse technological stack, further promoting transparency in AI training. 
    
\end{itemize}

\section{Related Works}
\label{sec:formatting}

MINT is grounded in the field known as Membership Inference Attacks (MIA), from which it derives its name. Despite fundamental differences, these two research areas also exhibit numerous similarities. Hence, we dive into the most noteworthy MIA work.

AI models are often trained on sensitive personal data \cite{melzi2025soft}, making them vulnerable to privacy leaks \cite{2026book-privacy}. This concept is the foundation of the Membership Inference Attacks (MIA) line. Essentially, MIA aims to extract this sensitive information through an adversarial approach. Shokri et al. \cite{shokri2017membership} pioneered this field, demonstrating MIA in standard datasets (CIFAR-10, UCI Adult, MNIST). Their approach relies on ``shadow models'' that emulate the functionality of the original model without direct access to it. Achieving this requires knowledge of the architecture and training methodology of the original model, along with limited samples from the original dataset and its aggregate statistics. Consequently, they created `shadow' training sets that mirror the original dataset and then trained shadow models to mimic the original model. This approach provides complete control over the training and non-training data of these models. With these shadow models and using the classification output vector for data on which they were trained and data on which they were not, they trained a binary classifier to identify them. They showed that the proper functioning of this binary classifier is highly dependent on overfitting.

Other approaches emerged from this line of research, following the same principle, but instead of training a binary classifier, they relied on the value of specific metrics and thresholds. For example, \cite{yeom2018privacy} is based on whether the loss value of an input data is above a threshold or not, or \cite{song2021systematic,salem2018ml}, which uses the prediction value.

Nasr et al. \cite{nasr2019comprehensive} introduced an important idea: the concept of Black-box and White-box terminologies. As mentioned earlier, they used the model's output vector (black-box); however, Nasr et al. proposed having access to all the information of the model (intermediate activations, gradients, weights), thus creating the so-called White-box scenario. The authors demonstrated that the utility of this White-box access is limited and does not offer improvements over the Black-box scenario. These results, coupled with the fact that White-box access in a scenario where you aim to attack a model is not a common situation, have led to limited literature in this area, and the existing literature does not achieve good results \cite{hu2022membership}.

More recently, Rezaei et al. \cite{rezaei2021difficulty} extensively examined the complexities associated with Membership Inference Attacks (MIA) and the inherent challenges of this task. Their principal discovery highlights that the practical performance of the task is considerably less favorable than that observed in state-of-the-art results. This discrepancy is mainly attributed to inadequacies in the evaluation methodology. Additionally, the researchers investigated the White-box scenario using activations, and again encountered challenges in achieving substantial performance improvements.

\section{Membership Inference Test}

Within the MINT framework, we position ourselves as model auditors who have access to the original model or insight into how the model processes the data. This scenario seamlessly aligns with the specific context of our work, which revolves around adapting to new legal frameworks and crafting innovative auditing tools. This distinguishes MINT from Membership Inference Attacks (MIA), where the conventional approach involves training ``shadow models" to emulate the behavior of the original model. In contrast, MINT capitalizes on direct access to the original model, enabling the application of our techniques without the need for shadow models. This distinction is game-changing and may lead to disparate detection results for tasks that are conceptually similar. 

Consider a Training Dataset ($\mathcal{D}$), an External Dataset ($\mathcal{E}$) and a collection of samples $d$ $(d \in \mathcal{D} \cup \mathcal{E})$. We assume a learned model ($M$) that is trained for a specific task (text generation, face recognition, etc.) using the dataset $\mathcal{D}$. For any input data record ($d$), the model ($M$) generates an outcome ($y$) based on $d$ and a set of parameters ($\textbf{w}$) learned during the training process, i.e., $y=M(d|\textbf{w})$. 

\begin{figure*}[t]
\centering
\includegraphics[width=0.85\linewidth]{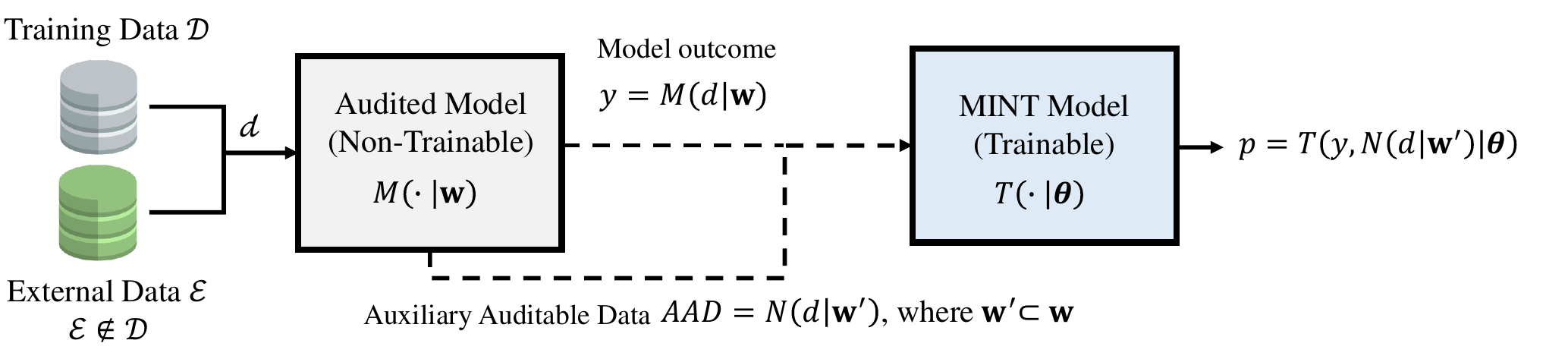} 
\caption{The Membership Inference Test (MINT) Model $T$ is trained to predict if specific data $d$ were used during the training process of an Audited Artificial Intelligence Model $M$ trained with a database $\mathcal{D}$. The input of the MINT Model is Auxiliary Auditable Data (e.g., activations maps for data samples $d$) and/or the model outcome obtained from $M$.}
\label{Block_Diagram_lite}
\end{figure*}

We assume that an authorized auditor has access to the model $M$. This access allows to obtain information about how $M$ processes the data $d$. This information comprises the generated result $y$ and, if possible, also some intermediate results (e.g., activation maps of specific layers in a Neural Network). These intermediate results $N(d|\textbf{w}')$ provide information about a limited set of parameters $\textbf{w}'$. We define these intermediate outcomes as Auxiliary Auditable Data.

The purpose of the Membership Inference Test (MINT) is to determine if data $d$ were used to train the model $M$. To this end, an authorized authority employs the Auxiliary Auditable Data to train an auditing model $T(\cdot|\theta)$ that is capable of predicting if a data sample $d$ belongs to the training dataset $\mathcal{D}$ or an External Dataset $\mathcal{E}$ ($\mathcal{E} \notin \mathcal{D}$). The key terms of the MINT method are defined below (see Fig.~\ref{Block_Diagram_lite}):

\begin{itemize}
    \item Audited Model ($M$): a learned model defined by an architecture and a set of parameters $\textbf{w}$.
    \item Training Data ($\mathcal{D}$): collection of data used to train $M$.
    \item External Data ($\mathcal{E}$): any data outside of the collection ($\mathcal{D}$).
    \item Model Outcome ($y = M(d|\textbf{w})$): final outcome of $M$ that results from processing input data $d$ 
    \item Auxiliary Auditable Data ($AAD = N(d|\textbf{w}')$): intermediate outcomes of $M$ that result from processing input data $d$ using a subset $\textbf{w}'$ of the parameters $\textbf{w}$. 
    \item MINT Model ($T$): a model defined by an architecture and a set of parameters $\theta$ trained using Auxiliary Auditable Data ($N(d|\textbf{w}')$) and/or Model Outcomes ($M(d|\textbf{w})$) of $M$ obtained from the two subsets of samples $\mathcal{D}$ and $\mathcal{E}$.
\end{itemize}

\section{MINT Applied to Text}

The MINT framework extends naturally beyond image modalities (where it was initially developed \cite{2024mint-factors,dealcala2025my}) to textual data \cite{mancera2026text-mint,mancera2026lora}, addressing growing concerns about the unauthorized use of text corpora in the training of large language models (LLMs). We present a MINT formulation adapted to LLMs fine-tuned with Low-Rank Adaptation (LoRA)~\cite{mancera2026lora}, a parameter-efficient fine-tuning method that specializes pre-trained models without updating all network parameters.

\subsection{LoRA Fine-Tuning and Membership Inference}

LoRA introduces a bottleneck structure consisting of two low-rank matrices $A \in \mathbb{R}^{d \times r}$ and $B \in \mathbb{R}^{r \times k}$ that approximate weight residuals for any frozen weight matrix $W_0 \in \mathbb{R}^{d \times k}$, where $r \ll \min(d, k)$. The adapted weight is defined as:
\begin{equation}
    W = W_0 + BA
\end{equation}
This low-rank factorization drastically reduces the number of trainable parameters while preserving model effectiveness. The distinct partition between static and trainable components makes LoRA-tuned architectures a particularly compelling subject for membership inference, as it may influence how sensitive information is distributed throughout the model.

\subsection{Feature Extraction and MINT Model for LLMs}

Given a base model $M$ fine-tuned with LoRA in $M_{\text{LoRA}}$, we define the MINT model as a parametric binary classifier $T(q(d) | \boldsymbol{\theta})$, where $q(d)$ is a fixed-dimensional feature representation that captures the model's internal and output responses to a given sample $d$. The MINT model predicts as follows:

\begin{equation}
    \hat{y} = T(q(d) \mid \boldsymbol{\theta}), \quad \hat{y} = \begin{cases} 1 & \text{if } d \in \mathcal{D} \\ 0 & \text{if } d \in \mathcal{E} \end{cases}
\end{equation}

The feature representation $q(d)$ integrates three complementary signal types:
\begin{itemize}
    \item Activation-based statistics: Mean, standard deviation, $\ell_2$ norm and average token-level $\ell_2$ norm extracted from the last three layers of the transformer of $M_{\text{LoRA}}$, which yields 12 features.
    \item Gradient-based sensitivity: The $\ell_2$ norm of the gradient with respect to the input embeddings, resulting in 1 feature.
    \item Output-level signals: Prediction loss and a confidence score (maximum predicted probability averaged across tokens), yielding 2 features.
\end{itemize}
This results in a 15-dimensional feature vector per sample. The MINT model is a fully connected neural network with two hidden layers (32 and 16 neurons), batch normalization, SiLU activations, and dropout, trained with binary cross-entropy loss using AdamW optimization.

This approach is particularly relevant in the context of AI governance: by requiring only limited and carefully selected internal signals from the model provider, it enables transparent compliance auditing without compromising the underlying model's integrity or performance.

\section{Experiments on Image Models}

The first experiments are with a popular face recognition model from the InsigthFace project \cite{insightface}. The model used ($M$ in Fig. \ref{Block_diagram_full}) is a ResNet-100 network \cite{han2017deep}, trained on the Glint360k database \cite{an2021partial} with CosFace loss function \cite{wang2018cosface}. This database comprises 17M images ($\mathcal{D}$ in Fig. \ref{Block_diagram_full}). 

We propose two different MINT model architectures:
\begin{enumerate}
    \item The first one called Vanilla MINT Model is based on a MLP: Three fully connected dense layers - 1) input-size neurons (varying depending on the Auxiliary Auditable Data used), 2) 64 neurons, and 3) 1 neuron. There is a dropout layer with a rate of 0.3 between them and an L1 regularizer with a value of 0.1.
    \item The second architecture named CNN MINT Model is based on a Convolutional Neural Network: Two convolutional layers with 64 and 128 filters, followed by two fully connected dense layers with sizes adapted to the output of the convolutional layers.
\end{enumerate}

We included the IJB-C \cite{IJB}, FDDB \cite{jain2010fddb}, GANDiffFace \cite{DEANDRESTAME2025103099} and Adience \cite{Adience} databases as external Data ($\mathcal{E}$) to train and test the MINT model $T$. As Auxiliary Auditable Data we used the activations \cite{2021_ICPR_InsideBias_Serna} of different layers in $M$. For the Vanilla MINT Model we take the activation maps at different depths of the network and extract the maximum value from each map, thus obtaining a vector. The size of the vector will depend on the number of filters in the selected convolutional layer. The CNN Mint Model analyzes activations directly without the need to vectorize them by taking the maximum of each map. The activation maps are introduced as-is into the CNN.

In Table~\ref{Table:AccForLayerNN}, we present the classification precision for the Vanilla MINT model. The table shows the results while varying the number of samples used to train the MINT model $T$ in columns and rows to represent the depth of the selected activation maps (the auditable adjunct data used). The ResNet-100 model consists of 4 major blocks, and we focus on the final convolutional layer of each block, which is reflected in the 4 layers depicted in the table (First to Fourth layers). The "output layer" is the output embedding of the model, and "all layers" means concatenating all the Conv Layers information. The classification accuracy varies depending on the available Auxiliary Auditable Data and the amount of data, revealing significantly better performance when using all available information, achieving up to 84\%. The best individual results are achieved with the auxiliary data closest to the network input and those closest to the output. intermediate layer results yield the poorest results.

\begin{table}
\centering
\resizebox{\columnwidth}{!}{%
\begin{tabular}{@{}lccc@{}}
\toprule
Auditable Data & \multicolumn{1}{l}{$1$K samples} & $50$K samples & $100$K samples \\ \midrule

Conv Layer \#1 & 0.62 & 0.80 & 0.80 \\
Conv Layer \#2 & 0.56 & 0.67 & 0.68 \\
Conv Layer \#3 & 0.56 & 0.58 & 0.59 \\
Conv Layer \#4 & 0.73 & 0.76 & 0.76 \\ 
Model Outcome & 0.67 & 0.78 & 0.78 \\
\textbf{All Conv Layers} & \textbf{0.76} & \textbf{0.82} & \textbf{0.84} \\ \bottomrule
\end{tabular}}
\caption{Classification accuracy for different Auxiliary Data and Model Outcome using the Vanilla MINT Model. The MINT model was trained with a variable number of samples ranging from $100K$ ($0.3\%$ of the total FR Model training set) to $1k$. }
\label{Table:AccForLayerNN}
\end{table}

In Table~\ref{Table:AccForLayerCNN}, we present the results for the CNN MINT Model. It is notable that there are no results for the Model Outcome in this table. This limitation arises from the incompatibility of applying the CNN architecture directly to the output vector. Similarly, the row representing the concatenation of information from the convolutional layers is absent due to the diverse resolutions of the activation maps, making concatenation unfeasible. This contrasts with the Vanilla MINT Model, where concatenation was viable after initial vectorization. For this architecture, optimal results are achieved with the layer closest to the model's input, diminishing as proximity to the model's output increases. Furthermore, this architecture yields the best results, reaching almost 90\% accuracy in detection compared to the 84\% obtained with the Vanilla MINT Model.

\begin{table}
\centering
\resizebox{\columnwidth}{!}{%
\begin{tabular}{@{}lccc@{}}
\toprule
Auditable Data & \multicolumn{1}{l}{$1$K samples} & $50$K samples & $100$K samples \\ \midrule

\textbf{Conv Layer \#1} & \textbf{0.88} & \textbf{0.89} & \textbf{0.89} \\
Conv Layer \#2 & 0.85 & 0.86 & 0.86 \\
Conv Layer \#3 & 0.68 & 0.71 & 0.75 \\
Conv Layer \#4 & 0.68 & 0.70 & 0.74 \\ \bottomrule
\end{tabular}}
\caption{Classification accuracy for different Auxiliary Data and Model Outcome using the CNN MINT Model. The MINT model was trained with a variable number of samples ranging from $100K$ ($0.3\%$ of the total FR Model training set) to $1k$. }
\label{Table:AccForLayerCNN}
\end{table}

\section{Experiments on Text Models}

We now evaluate the MINT framework applied to Large Language Models (LLMs) fine-tuned with Low-Rank Adaptation (LoRA)~\cite{mancera2026lora}. The audited models are obtained by fine-tuning a base LLM on approximately 13k samples, using a LoRA configuration with rank $r = 32$, scaling factor $\alpha = 32$, and dropout rate of 0.15, introducing trainable low-rank updates into selected transformer modules while preserving the original model parameters.

\subsection{Experimental Setup}

Experiments are conducted on three benchmark datasets that span different reasoning domains:

\begin{itemize}
    \item \textit{Medical-o1-reasoning-SFT}~\cite{chen2024huatuogpt}: A large-scale medical reasoning dataset containing verifiable question–answer pairs enriched with structured clinical reasoning trajectories for complex multi-step problem solving.
    \item \textit{CAMEL Math Reasoning}~\cite{li2023camel}: A synthetic multi-agent dataset focused on mathematical reasoning through structured role-playing dialogs, emphasizing step-by-step logical derivation.
    \item \textit{Maths \& College Reasoning}: A large-scale dataset covering college-level mathematical and academic reasoning, formatted as instruction–response pairs across algebra, calculus, probability, and statistics.
\end{itemize}

Four state-of-the-art LLMs are evaluated: Qwen3~\cite{bai2023qwen}, Phi-4~\cite{abdin2024phi4}, DeepSeek-V3~\cite{liu2024deepseek} and LLaMA~3.2-3B \cite{grattafiori2024llama3}. Together, they span a broad range of design choices in scale, training strategy, and architecture.

Six experimental configurations are evaluated, progressively incorporating deeper representations from $M_{\text{LoRA}}$: activation statistics from the antepenultimate (L1), penultimate (L2), and last (L3) transformer layers (each 7-dimensional); a gradient-based sensitivity feature alone (\textit{OnlyGrad}, 1-dimensional); a combination of gradient, loss, and confidence signals (\textit{GradProbLoss}, 3-dimensional); and a full integration of all three layers with gradient and output features (\textit{All3Layers}, 15-dimensional). All configurations are compared with three standard MIA baselines: MinK~\cite{shi2023mink}, MinK++~\cite{zhang2024minkpp} and Loss~\cite{liu2022membership}. Performance is reported as AUC.

\subsection{Results and Analysis}

Table~\ref{tab:text_results} summarizes the AUC scores under all experimental conditions. We discuss the main findings below.

\textit{Layer depth progressively improves membership detection,} while performance improves consistently as deeper transformer layers are used. Moving from L1 to L3 yields systematic gains across all models and datasets — for example, on Medical-o1, Qwen3 improves from 0.8306 to 0.9272 and DeepSeek-V3 from 0.9697 to 0.9846. This confirms that later layers encode stronger membership-related signals, likely due to increased task-specific memorization during fine-tuning. In particular, even L1 alone surpasses all output-only baselines in most configurations.

\textit{Gradient-only features are insufficient for robust inference,} so the \textit{OnlyGrad} configuration consistently produces the lowest performance across all settings (e.g., AUC = 0.5927 for LLaMA~3.2-3B on Medical-o1), suggesting that gradient sensitivity alone is too noisy to reliably distinguish training from non-training samples. However, increasing it with prediction loss and confidence score (\textit{GradProbLoss}) leads to substantial recovery — DeepSeek-V3 in Medical-o1 improves from 0.6683 to 0.9435 — highlighting the complementary nature of these signals.

\textit{Multilayer fusion achieves the best overall performance,} with \textit{the All3Layers} configuration consistently achieving the highest AUC scores, reaching 0.9901 for DeepSeek-V3 in both CAMEL and Math \& College Reasoning, and 0.9857 for Qwen3 on the same dataset. Even for smaller models such as LLaMA~3.2-3B, All3Layers achieves competitive results (e.g. AUC = 0.9792 in Math \& College Reasoning), confirming that membership information is distributed across multiple model levels and that leveraging heterogeneous multi-layer signals is essential for robust inference.

\textit{MINT consistently outperforms MIA baselines.} All layer-based MINT variants outperform MinK, MinK++, and Loss across datasets and models, demonstrating that exploiting internal representations — even with limited model access — is more informative than relying solely on output-level signals.

\textit{The model scale influences privacy leakage;} larger models, such as DeepSeek-V3, achieve higher and more stable AUC scores, suggesting stronger memorization effects. LLaMA~3.2-3B shows the most variable performance, while Qwen3 and Phi-4 occupy an intermediate regime, indicating that scale and architecture play a meaningful role in determining how detectable membership information is.

\begin{table}[t]
\caption{Performance comparison (AUC) of MINT-based variants and baseline MIA methods across three datasets and four LoRA fine-tuned LLMs. Layer configurations: antepenultimate (L1), penultimate (L2), and last layer (L3). Best result per model and dataset in \textbf{bold}.}
\label{tab:text_results}
\centering
\resizebox{\columnwidth}{!}{%
\begin{tabular}{llcccc}
\hline
\textbf{Dataset} & \textbf{Method} & \textbf{Qwen3} & \textbf{Phi-4} & \textbf{DeepSeek-V3} & \textbf{LLaMA 3.2-3B} \\
\hline
\multirow{9}{*}{\shortstack[l]{Medical-o1\\reasoning-SFT}}
 & MinK              & 0.7690 & 0.9440 & 0.8840 & 0.7770 \\
 & MinK++            & 0.7780 & 0.9590 & 0.8625 & 0.7812 \\
 & Loss              & 0.7508 & 0.9390 & 0.8840 & 0.8000 \\
 & L1                & 0.8306 & 0.9800 & 0.9697 & 0.8811 \\
 & L2                & 0.9179 & 0.9788 & 0.9840 & 0.9202 \\
 & L3                & 0.9272 & 0.9811 & 0.9846 & 0.9210 \\
 & OnlyGrad          & 0.6706 & 0.7132 & 0.6683 & 0.5927 \\
 & GradProbLoss      & 0.7920 & 0.6323 & 0.9435 & 0.8394 \\
 & \textbf{All3Layers} & \textbf{0.9302} & \textbf{0.9819} & \textbf{0.9851} & \textbf{0.9240} \\
\hline
\multirow{9}{*}{\shortstack[l]{CAMEL Math\\Reasoning}}
 & MinK              & 0.8720 & 0.9160 & 0.9050 & 0.8090 \\
 & MinK++            & 0.7823 & 0.8730 & 0.8465 & 0.7985 \\
 & Loss              & 0.8838 & 0.9160 & 0.9051 & 0.8090 \\
 & L1                & 0.9294 & 0.9533 & 0.9841 & 0.8948 \\
 & L2                & 0.9429 & 0.9496 & 0.9865 & 0.9591 \\
 & L3                & 0.9485 & 0.9480 & 0.9843 & 0.9630 \\
 & OnlyGrad          & 0.7073 & 0.8047 & 0.7895 & 0.7536 \\
 & GradProbLoss      & 0.9201 & 0.9400 & 0.9254 & 0.8639 \\
 & \textbf{All3Layers} & \textbf{0.9514} & \textbf{0.9578} & \textbf{0.9901} & \textbf{0.9645} \\
\hline
\multirow{9}{*}{\shortstack[l]{Maths \& College\\Reasoning}}
 & MinK              & 0.9440 & 0.9021 & 0.9840 & 0.8490 \\
 & MinK++            & 0.9590 & 0.8464 & 0.8945 & 0.7954 \\
 & Loss              & 0.9380 & 0.9210 & 0.9610 & 0.7850 \\
 & L1                & 0.9848 & 0.9521 & 0.9810 & 0.9747 \\
 & L2                & 0.9855 & 0.9801 & 0.9845 & 0.9743 \\
 & L3                & 0.9810 & 0.9821 & 0.9866 & 0.9726 \\
 & OnlyGrad          & 0.6548 & 0.8424 & 0.7462 & 0.8420 \\
 & GradProbLoss      & 0.9779 & 0.9398 & 0.9785 & 0.9627 \\
 & \textbf{All3Layers} & \textbf{0.9857} & \textbf{0.9845} & \textbf{0.9901} & \textbf{0.9792} \\
\hline
\end{tabular}%
}
\end{table}



\section{Membership Inference Test: \\Works perspective}

Contextualizing our results is crucial; however, we focus our comparison on MIA as the primary reference. In doing so, it is imperative to recognize that MIA differs in initial conditions, which can lead to divergent outcomes.

Even within the MIA framework, there is a dearth of studies addressing the detection of training samples for facial recognition, excluding direct comparisons. Nevertheless, we can provide contextualization through other MIA works that tackle similar tasks. Throughout the state-of-the-art \cite{hu2022membership, MIAAgainstLiu}, numerous experiments and advances have been conducted, all within what we term ``toy examples," utilizing simplistic datasets such as CIFAR-10, CINIC-10, or CIFAR-100, and under specific overfitting conditions. The accuracy values achieved in the state-of-the-art for these toy examples hover around 65\%, 73\%, and 80\%, respectively. In contrast, our model operates in a competitive real-world facial recognition environment where identifying training data is highly critical. Furthermore, our results reach up to 90\%, markedly exceeding the state-of-the-art works. Contrary to previous state-of-the-art conclusions, we have successfully enhanced the results by utilizing model activations in a white-box environment.

\section{Membership Inference Test: Demonstrator}

The main objective of this work is to promote our web platform \footref{note:plataforma}, where we make MINT available to users. On this platform, citizens can upload their personal data and receive a report indicating the probabilities that these data were used to train an AI model from a list of available models. In this MINT Demo 2 version, we include both image models \cite{dealcala2025my} and text models \cite{mancera2026text-mint} developed through various technologies (MINT \cite{2024mint-factors,dealcala2025my}, aMINT \cite{dealcala2025active}, or gMINT \cite{dealcala2025gmint}). This platform opens up new opportunities for research and encourages the development of tools, standards, and protocols to comply with the new regulations.

\begin{figure}[t]
\centering
\includegraphics[width=\linewidth]{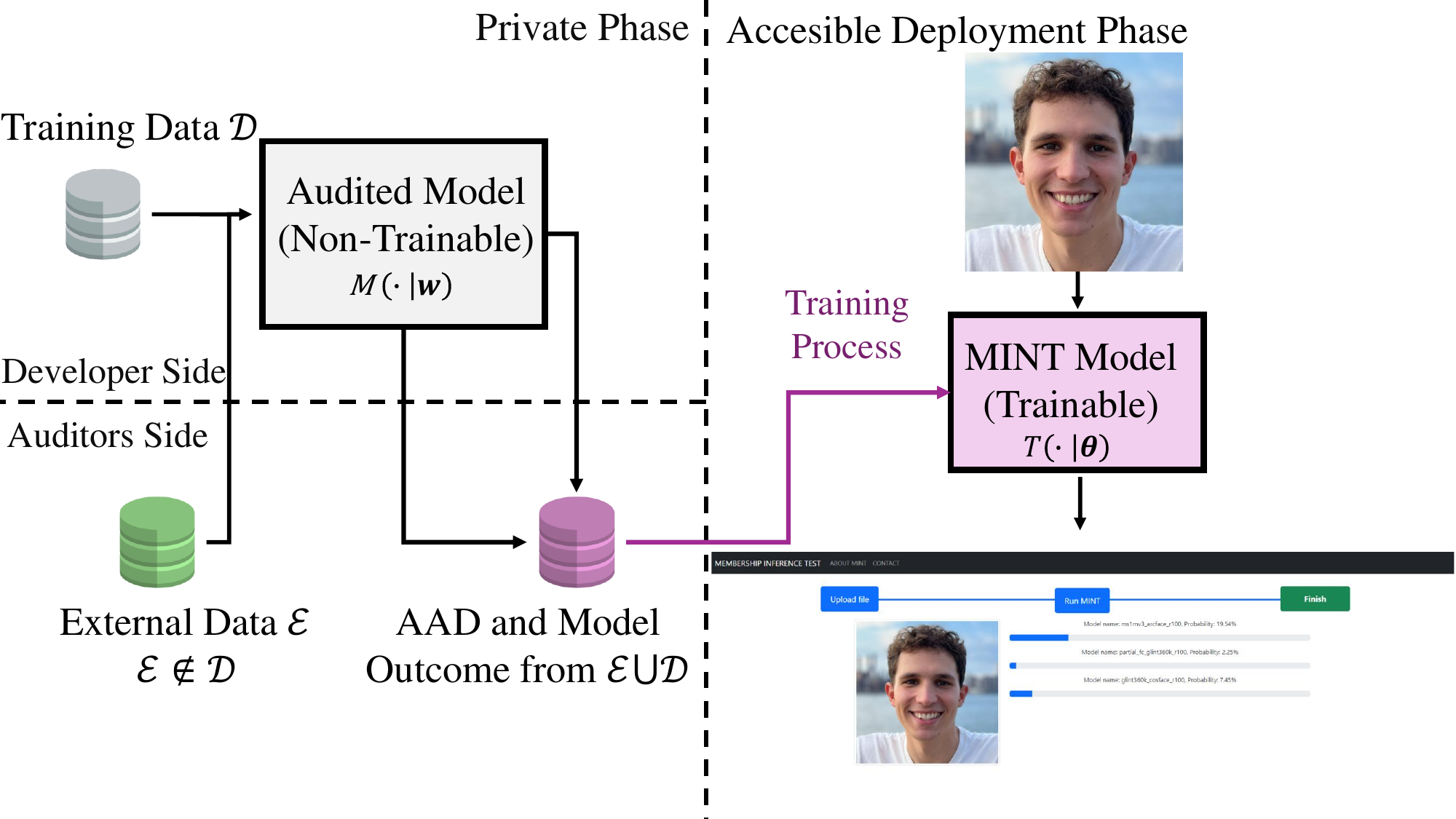} 
\caption{Block diagram of the MINT platform operation.} 
\label{Block_diagram_platf}
\end{figure}

In Fig.~\ref{Block_diagram_platf}, we can observe the flow of the web operation. Initially, the developer of the Audited Model $M$ contacts our API. This developer provides us with the information necessary to train our MINT mode, including the Audited Model $M$ and some training samples $\mathcal{D}$. In our end, we possess a private set of samples that have not been used in the training of the model $\mathcal{E}$. With these two sets, we obtain the $AAD$ and/or the outcome of the model required to train our MINT Model. The Audited Model is never made public on the platform; we have exclusive access to it solely for training the MINT Model, which is subsequently made public and available to users. It is important to emphasize that the private set consists of samples that are not used for training and are utilized to train the MINT Model. However, this MINT Model is capable of generalizing for any sample not used for training, not just for the private set.

\section{Conclusions}

We introduce MINT Demo 2, a platform that aims to empower users to determine the data utilized in the training of a model. With this platform, our goal is to provide tools to ensure compliance with new regulations and to promote transparent and responsible AI practices. Our goal is to encourage widespread participation from the scientific community and facilitate the audit of as many models as possible.

\section{Acknowledgment}

Support by project M2RAI (PID2024-160053OB-I00, MICIU/FEDER) and Cátedra ENIA UAM-VERIDAS en IA Responsable (NextGenerationEU PRTR TSI-100927-2023-2). Mancera is supported by FPI-PRE2022-104499 MICINN/FEDER and DeAlcala is supported by FPU21/05785 MIU. Work conducted within the ELLIS Unit Madrid. 

\bibliographystyle{IEEEtran}
\bibliography{main}

\end{document}